\documentclass[conference]{IEEEtran}
\usepackage{cite}
\usepackage{amsmath,amssymb,amsfonts}
\usepackage{algorithmic}
\usepackage{graphicx}
\usepackage{caption}
\usepackage{textcomp}
\usepackage{xcolor}
\usepackage{hyperref}
\usepackage{comment}
\usepackage{balance}
\hypersetup{
    colorlinks = true,
    linkcolor = blue,
    citecolor = green,
    urlcolor = cyan
}

\def\BibTeX{{\rm B\kern-.05em{\sc i\kern-.025em b}\kern-.08em
    T\kern-.1667em\lower.7ex\hbox{E}\kern-.125emX}}
    
\begin{document}

\title{A Survey on Mamba Architecture for Vision Applications}

\author{\IEEEauthorblockN{Fady Ibrahim}
\IEEEauthorblockA{\textit{Department of Computer Science} \\
\textit{Toronto Metropolitan University}\\
Toronto, Canada \\
f1ibrahim@torontomu.ca}
%https://orcid.org/0009-0004-6534-598X}

\and
\IEEEauthorblockN{Guangjun Liu}
\IEEEauthorblockA{\textit{Department of Aerospace Engineering} \\
\textit{Toronto Metropolitan University}\\
Toronto, Canada \\
gjliu@torontomu.ca}

\and
\IEEEauthorblockN{Guanghui Wang}
\IEEEauthorblockA{\textit{Department of Computer Science} \\
\textit{Toronto Metropolitan University}\\
Toronto, Canada \\
wangcs@torontomu.ca}
}

\maketitle

\begin{abstract}
Transformers have become foundational for visual tasks such as object detection, semantic segmentation, and video understanding, but their quadratic complexity in attention mechanisms presents scalability challenges. To address these limitations, the Mamba architecture utilizes state-space models (SSMs) for linear scalability, efficient processing, and improved contextual awareness.
This paper investigates Mamba architecture for visual domain applications and its recent advancements, including Vision Mamba (ViM) and VideoMamba, which introduce bidirectional scanning, selective scanning mechanisms, and spatiotemporal processing to enhance image and video understanding. Architectural innovations like position embeddings, cross-scan modules, and hierarchical designs further optimize the Mamba framework for global and local feature extraction. These advancements position Mamba as a promising architecture in computer vision research and applications.
\end{abstract}
\vspace{12pt}

\section{Introduction}\label{sec:intro}
Deep learning has revolutionized artificial intelligence (AI), and has been instrumental in advancing computer vision, significantly enhancing machines' ability to interpret and understand visual information from the world. By harnessing the power of neural networks, particularly Convolutional Neural Networks (CNNs) \cite{li2022longseqcnn, lioutas2020timekernelcnn} and, more recently, Vision Transformers (ViTs) \cite{zhang2025depth, vaswani2017attention, dosovitskiy2020vit}, deep learning has enabled the development of complex, highly accurate models for tasks such as image classification \cite{alapati2024predicting}, object detection \cite{zhang2023aphid}, semantic segmentation \cite{rahman2024new, xiao2023edge}, and spatio-temporal video understanding. \cite{he2016deepresimagerec, bertasius2021space}.
Transformers, originally popularized in natural language processing (NLP) \cite{vaswani2017attention}, have shown significant potential in vision tasks. Architectures such as Vision Transformers (ViTs) \cite{dosovitskiy2020vit} and hybrid models \cite{child2019generating, ali2021xcit} leverage their ability to capture complex dependencies, delivering state-of-the-art results.

At the core of these advances are models with massive parameter counts capable of capturing intricate patterns in data. Transformers despite their success, face challenges due to the quadratic complexity of attention calculations, resulting in time-consuming inference. Mamba, a novel architecture inspired by state-space models (SSMs), presents a promising alternative for building foundation models \cite{gu2023mamba}. With near-linear scalability in sequence length and hardware-efficient processing, Mamba has been recently adapted to computer vision tasks, including image and video processing. Models like Vision Mamba (ViM) \cite{zhu2024vision} and VideoMamba \cite{li2025videomamba} demonstrate Mamba's ability to handle long-range dependencies while addressing the computational and memory bottlenecks faced by Transformer-based models in the visual domain.

This paper offers a comprehensive review of Mamba architectures applied to visual tasks, highlighting their similarities, differences, and effectiveness in addressing various challenges. 
Our survey exclusively focuses on Mamba's role in visual tasks, setting it apart from broader Mamba surveys that cover applications in text, recommendation systems, and medical imaging \cite{qu2024mambasurvey, patro2024surveylongseqmamba, yue2024medmamba, bansal2024mambamedimagesurvey}. 
While previous works like \cite{xu2024vimsurvey, zhang2024vismambasurvey} examine Mamba in vision, this paper uniquely focuses on ViM and VideoMamba as core models. We assess architectural enhancements, including position embeddings, cross-scan modules, and hierarchical designs, to provide new insights into optimizing Mamba for both local and global feature extraction. Additionally, we offer a comparative performance analysis across image classification, semantic segmentation, and object detection, giving practical recommendations for selecting the best architectures for various vision tasks. This task-specific comparison sets our work apart from general Mamba surveys and provides valuable guidance for vision-based research.
%Although previous work such as \cite{xu2024vimsurvey, zhang2024vismambasurvey} also examine Mamba’s use in vision, this paper uniquely anchors its analysis around Vision Mamba (ViM) and VideoMamba as foundational models. By evaluating architectural enhancements such as position embeddings, cross-scan modules, and hierarchical designs, we offer a fresh perspective on optimizing Mamba for local and global feature extraction. This survey also provides a comparative performance analysis on standard datasets in image classification, semantic segmentation, and object detection, making practical recommendations for selecting the best architectures for different vision tasks. This structured architectural comparison based on task-specific applications distinguishes our work from general Mamba surveys and offers valuable insights for researchers focusing on vision-based applications.

The rest of this survey is organized as follows: Section~\ref{sec:prelim} provides an introductory overview of the building blocks of ViM and the theoretical foundations of SSMs. Section~\ref{sec:visund} compares ViM and Video Mamba. Section~\ref{sec:keyarch} reviews recent advancements in Mamba architectures. Section~\ref{sec:results} presents a performance comparison and analysis of the novel architectures on common datasets. %Section~\ref{sec:discuss} discusses our recommendation on which architectures to use for different tasks. 
Section~\ref{sec:chall} discusses the challenges associated with ViM and its applications. Finally, the paper is concluded in Section \ref{sec:concl}.
    
\begin{comment}
The contributions of this paper are as follows:
\begin{enumerate}
    \item \textbf{Preliminary} (Section~\ref{sec:prelim}): An introductory overview of the building blocks of Vision based Mamba models, the theoretical calculations behind how State Space Models function, and the progression from Vision Mamba utilized for image inputs to VideoMamba for video data.
    \item \textbf{Visual Understanding: Vision Mamba Vs. VideoMamba} (Section ~\ref{sec:visund}): Comparison of Vision Mamba and Video Mamba from the perspectives of block design, scanning mode, and memory management.
    \item \textbf{Key Architectures} (Section~\ref{sec:keyarch}): Overview of the recent advancements in Mamba architectures and how each addresses different tasks using their own novel contributions and modalities.
    \item \textbf{Results: Comparison} (Section~\ref{sec:results}): Tabulated results comparing the novel architectures performance against common datasets.
    \item \textbf{Challenges and Future Research} (Section~\ref{sec:chall}): Examination of the challenges associated with applying Mamba for visual applications and tasks. As well as an exploration of prospective future research focusing on novel SSM architectures and more efficient algorithms.
\end{enumerate}
\end{comment}

\section{Preliminary}\label{sec:prelim}

%This section provides an overview of the building blocks of vision-based Mamba models and details how the Mamba architecture was adapted for visual data, enabling it to process sequences of image and video patches.

\subsection{Mamba Block}

\begin{figure}[t]
\centering
\includegraphics[width=\linewidth]{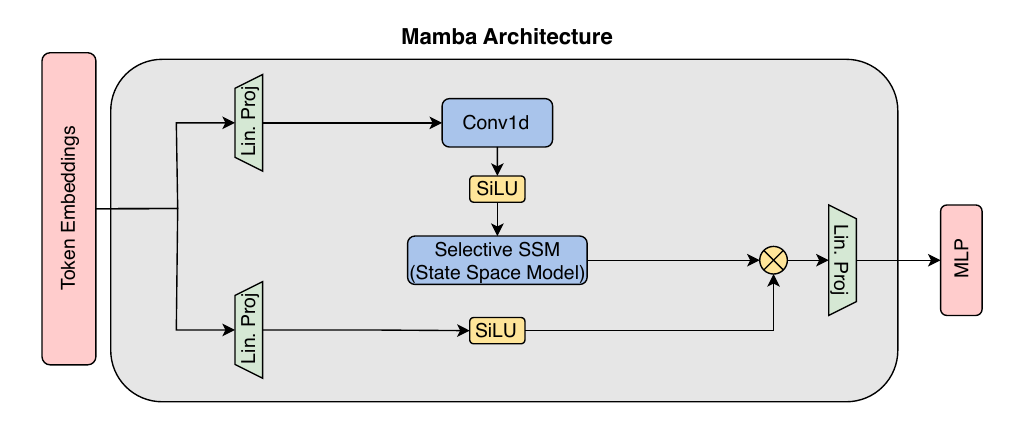}
\caption{Mamba Block Architecture with selective state space models.}%, originally proposed for large language models with the ability to remember or ignore inputs depending on their content.}
\label{fig1}
\end{figure}

Mamba enhances the context-aware capabilities of traditional SSMs by making its parameters functions of the input. Fig.~\ref{fig1} shows the original Mamba architecture block intended for 1D sequential data such as language tokens. SSMs map a 1D function or sequence $x(t) \in \mathbb{R}$ to an output $y(t) \in \mathbb{R}$ through a hidden state $h(t) \in \mathbb{R}^N$.
Using parameters 
$A \in \mathbb{R}^{N \times N}$$, B \in \mathbb{R}^{N \times 1}$$, $$C \in \mathbb{R}^{1 \times N}$, the continuous system is defined by

\begin{equation}
\begin{aligned}
\frac{dh(t)}{dt} = Ah(t) + Bx(t), \quad y(t) = Ch(t)
\end{aligned}
\label{eq:ssm}
\end{equation}

For the discrete sequences defined by the input \( \mathbf{x} = (x_0, x_1, \dots) \in \mathbb{R}^L \), the continuous system parameters in Eq.~\eqref{eq:ssm} must be discretized using a step size \( \Delta \). Zero Order Hold (ZOH)\cite{zhang2007comparediscretize} is used as the simplest and most reliable discretization method in\cite{zhu2024vision, li2025videomamba, liu2024vmambavisualstatespace}. Using step size \( \Delta \), the continuous system parameters \( A, B \) are converted to \( \bar{A}, \bar{B} \) as

\begin{equation}
\begin{aligned}
\bar{A} &= \exp(\Delta A), \\
\bar{B} &= (\Delta A)^{-1} (\exp(\Delta A) - I) \cdot \Delta B.
\end{aligned}
\label{eq:zoh}
\end{equation}
and the discretized system equation can be represented as

\begin{equation}
\begin{aligned}
h_t = \bar{A}h_{t-1} + \bar{B}x_t, \quad y_t = Ch_t
\end{aligned}
\label{eq:ssmdisc}
\end{equation}

\subsection{Selective Scanning Mechanism}
Traditional SSMs struggle with maintaining context awareness in dense text and data. Mamba introduces a selection mechanism that dynamically adapts to input data, filtering irrelevant information while retaining relevant data indefinitely. Mamba's innovation lies in the Selective Scan Mechanism (S6) \cite{gu2023mamba}, making SSM parameters $B$, $C$, $\Delta$ functions of the input as $S_B(x), S_C(x), S_\Delta(x)$. This is achieved through a linear transformation 
$$B, C, \Delta = \text{Linear}(x)$$ 
where $B, C, \Delta \in \mathbb{R}^{B \times L \times N}$, $L$ denotes the input sequence length, $B$ is the batch size, and $D$ denotes the number of channels.

%\vspace{12pt}
The original Mamba block \cite{gu2023mamba} introduced a hardware-aware algorithm to efficiently compute the selective SSM parameters.
These parameters are used in SSM equations Eq.~\eqref{eq:ssm} and Eq.~\eqref{eq:ssmdisc} to compute the latent hidden state and output. This mechanism allows Mamba to selectively retain or discard information based on the relevance of the input sequence.
\begin{equation}
h_t = \bar{A}h_{t-1} + \bar{B}x_t, \quad y_t = S_Ch_t
\label{eq:selscanssmdisc}
\end{equation}
where
\begin{equation}
\begin{aligned}
\bar{A} &= \exp(S_\Delta A)\\ 
\bar{B} &= (S_\Delta A)^{-1} (\exp(S_\Delta A) - I) \cdot S_\Delta S_B.
\end{aligned}
\label{eq:selscanzoh}
\end{equation}

\color{black}
\subsection{Vision Mamba (ViM) Block}
\begin{figure}[t]
\centering
\includegraphics[width=\linewidth]{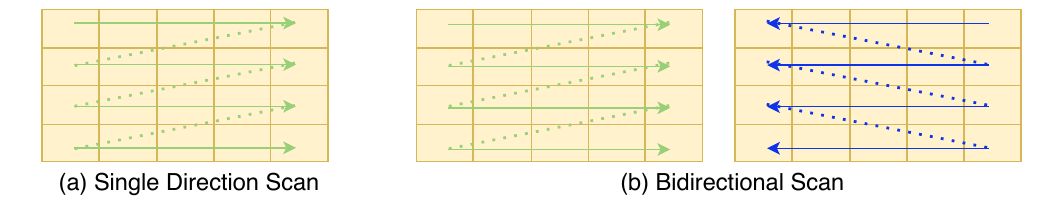}
\caption{Single direction scanning mechanism on patch embeddings shown in (a) follows the original Mamba architecture. Bidirectional Selective Scanning (b) introduces as a novel contribution in ViM.}
\label{fig2}
\end{figure}

Vision Mamba extends the Mamba architecture to address the complexities of visual data. It tackles position sensitivity by incorporating position embeddings, similar to those used in ViT \cite{dosovitskiy2020vit}. To capture global context, ViM employs bidirectional SSMs, as illustrated in Fig.~\ref{fig2}, allowing each sequence element to access information from both preceding and succeeding elements \cite{zhu2024vision}.

\subsection{Bidirectional/Spatiotemporal Scanning}
\begin{figure}[t]
\centering
\includegraphics[width=\linewidth]{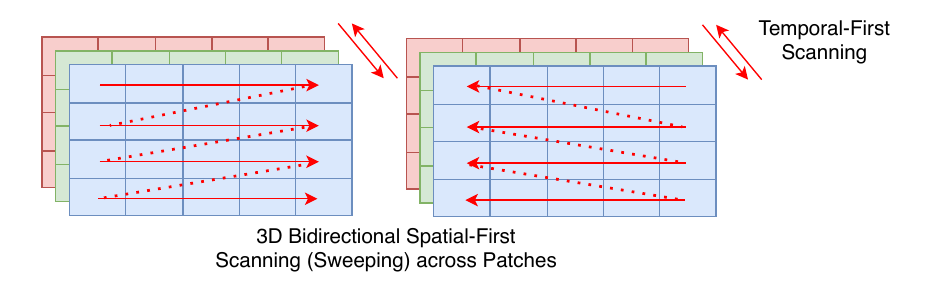}
\caption{3D Spatiotemporal Bidirectional Scanning. VideoMamba \cite{li2025videomamba} introduces an enhanced scanning mechanism for 3D input data to combine spatial data with temporal data.}
\label{fig3}
\end{figure}

The original Mamba block's unidirectional scanning suits causal sequences. However, visual data is non-causal. ViM employs bidirectional scanning, processing sequences forward and backward to capture dependencies effectively, as shown in Fig.~\ref{fig2}. For video understanding, spatiotemporal scanning extends this concept. VideoMamba employs 3D bidirectional blocks to capture spatial and temporal dependencies, as shown in Fig.~\ref{fig3}. In addition to these strategies, researchers have explored various spatiotemporal methods, such as spatial-first versus temporal-first scanning, as well as sweeping scan, continuous scan, and local scan techniques, to further enhance ViM's performance and efficiency. An overview of how these methods are utilized in key architectures is presented in Section IV.

\section{Vision Mamba vs. VideoMamba}\label{sec:visund}

This section compares the Vision Mamba block with the more complex VideoMamba model, focusing on block design, scanning mode, and memory management.
\begin{figure*}[t]
\centering
\includegraphics[width=0.97\textwidth,height=\textheight,keepaspectratio]{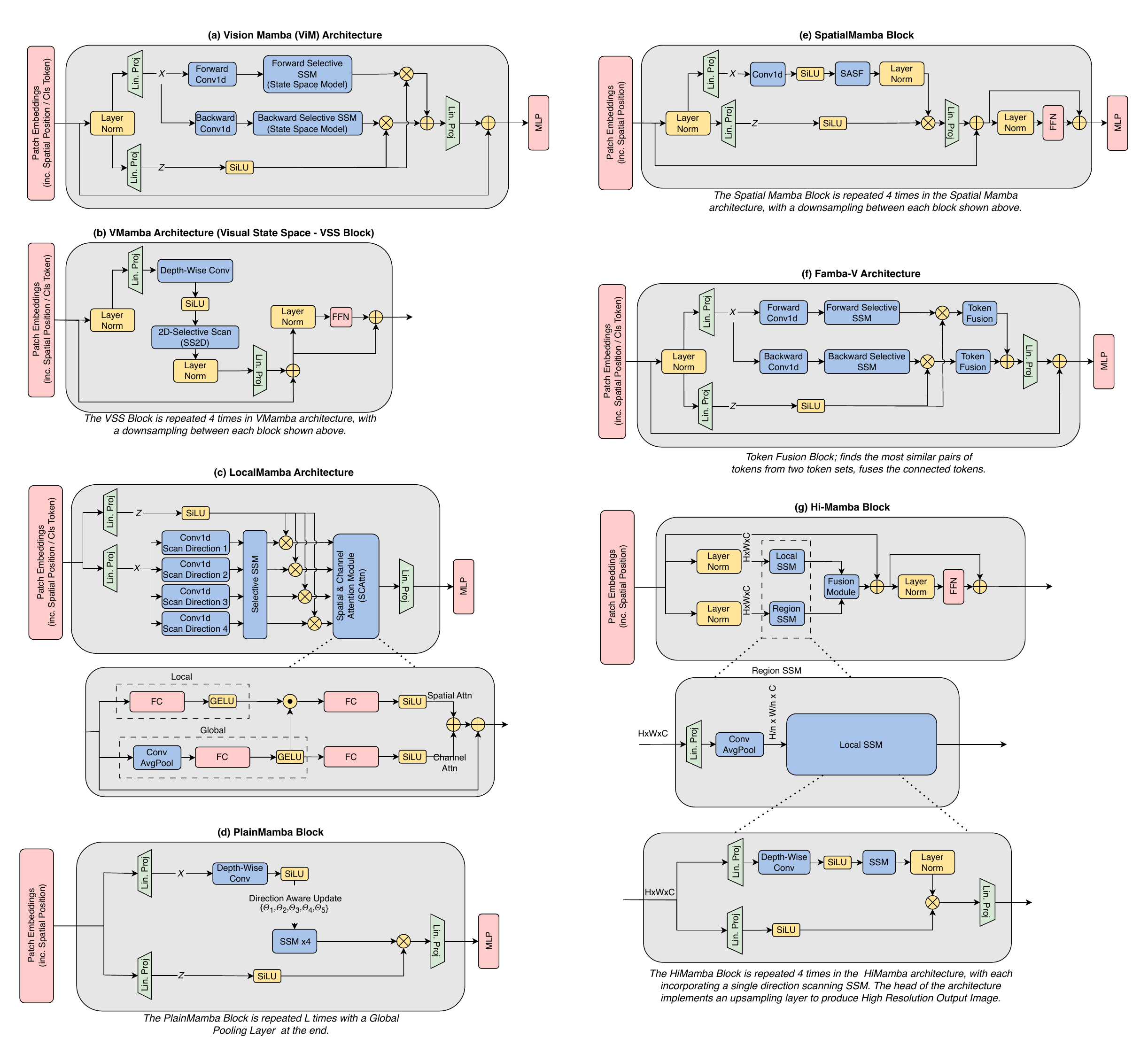}
\caption{Comparison of Mamba architectures for 2D image applications; (a) Vision Mamba, (b) Vmamba, (c) LocalMamba is a set of multidirectional blocks that can be applied to other Mamba architectures, (d) PlainMamba, (e) SpatialMamba, (f) Famba-V introduces the concept of token fusion that can be applied to other Mamba architectures, (g) Hi-Mamba, used to extract high-resolution image features and for high-resolution image restoration.}
\label{figfull}
\end{figure*}

\subsection{Vision Mamba (ViM)}
The ViM block as shown in Fig.~\ref{figfull}(a) is an adaptation of the Mamba block designed to process visual data using \textbf{bidirectional scanning/sequence modeling}. Unlike the original unidirectional Mamba block the ViM block processes flattened visual sequences with simultaneous forward and backward SSMs to capture spatial context. 
The input sequence is generated by transforming the 2D image into flattened 2D image patches with the addition of both position embeddings and class tokens; the latter represents the whole patch sequence.

The normalized sequence input is linearly projected to produce two transformed representations: $x$ (used for main processing) and $z$ (used later for gating).
%\begin{itemize}
%    \item \( x \) (used for main processing).
%    \item \( z \) (used later for gating).
%\end{itemize} 
The \( x \) transformed representation is processed in both \textbf{forward} and \textbf{backward} directions (similar to bidirectional RNNs or Transformers).
For each direction:
\begin{itemize}
    \item A \textbf{1D convolution} is applied to \( x \), followed by a \textbf{SiLU activation}, producing \( x'_{o} \); represented by the "Forward Conv1d and Backward Conv1d" blocks in Fig.~\ref{figfull}(a).
    \item Within the SSM blocks, three separate \textbf{linear projections} transform \( x'_{o} \) into:
    $
    B_o, C_o, \Delta_o
    $.
    \item The scaling factor \( \Delta_o \) is applied to transform \(A_o\), \(B_o\):
    
    \begin{equation}
    \begin{aligned}
    \bar{A_o} &= \exp(\Delta_o A_o), \\
    \bar{B_o} &= (\Delta_o A_o)^{-1} (\exp(\Delta_o A_o) - I) \cdot \Delta_o B_o.
    \end{aligned}
    \label{eq:zoh_vim}
    \end{equation}
    
    \item The SSM then computes forward-scanned (\( y_{\text{forward}} \) and backward-scanned \( y_{\text{backward}} \)) output features, which are gated and averaged to produce the output. 
\end{itemize}

%\vspace{12pt}

The outputs (\( y_{\text{forward}} \) and \( y_{\text{backward}} \)) are modulated using \textbf{SiLU activations} applied to \( z \); This acts as a \textbf{gating mechanism}, controlling the contribution of each direction.  
The gated outputs are \textbf{added together} and passed through a final \textbf{linear transformation}, where a \textbf{residual connection} (adding back the original normalized sequence input ensures that useful original information is retained.

This \textbf{bidirectional processing} is a key difference from the original Mamba block, allowing the ViM block to consider the context from both past and future tokens, as image data is inherently non-causal, and a token in a future sequence segment may have significant relevance to the context of the currently assessed image token. The ViM block incorporates \textbf{position embeddings} to preserve spatial information, addressing the original Mamba's lack of spatial awareness as it was designed for 1D sequences. Additionally, parameters are shared for both forward and backward scanning directions within the ViM block. By incorporating bidirectional SSMs and position embeddings, ViM and all of its variants aim to achieve comparable modeling power to Transformers while maintaining a linear memory complexity and efficiently compressing visual representations, making it suitable for various vision tasks. While effective, ViM faces the following unique limitations:

\begin{itemize}
    \item Presence of Artifacts: One study reveals that ViM feature maps can contain artifacts, potentially impacting performance \cite{wang2024mambareg}. These artifacts necessitate further refinements to the architecture for optimal results.
    \item Mitigating Artifacts with Mamba®: To address artifacts, researchers propose Mamba®, which strategically incorporates "register tokens" evenly distributed throughout the token sequence. These tokens are then concatenated at the end of the network to generate a comprehensive image representation. Mamba® effectively reduces artifacts, sharpens the model's focus on semantically relevant image areas, and surpasses previous models in accuracy and scalability \cite{wang2024mambareg}.
    \item Limited global and local representations concurrently: Researchers have noted that the unidirectional nature of traditional Vision Mamba architectures can limit their ability to effectively capture global representations, potentially reducing performance in tasks requiring comprehensive spatial context. \cite{shi2024vssd, huang2024localmamba}.
\end{itemize}

There have been several ViM variations designed to address these specific challenges and more to enhance performance across various tasks:
\begin{itemize}
    \item Vmamba: This model introduces the Cross-Scan Module (CSM) to specifically address the two-dimensional nature of images. By considering spatial relationships on a 2D plane as other models have also done \cite{baty2024mamba2d}, Vmamba aims to improve the efficiency of visual processing \cite{liu2024vmambavisualstatespace}.
    \item LocalMamba and EfficientVMamba: These variations focus on optimizing scanning strategies within the Mamba architecture. LocalMamba employs window scanning and dynamic scanning directions \cite{huang2024localmamba}, while EfficientVMamba integrates atrous-based selective scanning and dual-pathway modules for efficient global and local feature extraction \cite{lee2024efficientvim}.
    \item VSSD Mamba: This model addresses the fundamental challenge of adapting the causal nature of traditional SSD/SSM architectures to inherently non-causal vision tasks \cite{dao2024mamba2}. It overcomes this by disregarding the magnitude of interactions between hidden states and tokens, retaining only their relative weights. This technique, along with multi-scan strategies, enables the non-causal integration of scanning results, significantly improving performance in vision applications \cite{shi2024vssd}.
\end{itemize}

\subsection{Video Mamba}

VideoMamba is designed specifically for video understanding, addressing the challenges of local redundancy and global dependencies in video data. It extends the ViM block for 3D video sequences, incorporating a bidirectional Mamba block, and employs various scanning methods, such as Spatial-First, Temporal-First, and Spatiotemporal scans, to effectively process spatiotemporal information, as illustrated in Fig.~\ref{fig3}. The model incorporates a linear-complexity operator to handle long-term dependencies efficiently and includes self-distillation, where a smaller model acts as a teacher, guiding the training of the larger model to improve scalability.

VideoMamba \cite{li2025videomamba} extracts key frames from videos and processes them as continuous input sequences. However, its performance on video benchmarks falls short of transformer-based methods, indicating room for improvement \cite{arnab2021vivit}. Building upon VideoMamba, \cite{lu2024videomambapro} addresses the limitations of ``historical decay" and ``element contradiction". Masked backward computation mitigates the historical decay issue, allowing the network to better handle historical tokens. Extensions of VideoMamba models \cite{lu2024videomambapro, chen2024video} implement residual connections to refine the representation of similarity between tokens. These enhancements significantly boost VideoMamba's performance on video action recognition benchmarks, establishing it as a strong competitor to transformer models.

While both models leverage bidirectional SSMs, VideoMamba extends this capability to spatiotemporal processing and incorporates distinct scanning methods, masked modeling strategies, and self-distillation for enhanced performance.

\begin{figure}[t]
\centering
\includegraphics[width=0.95\linewidth]{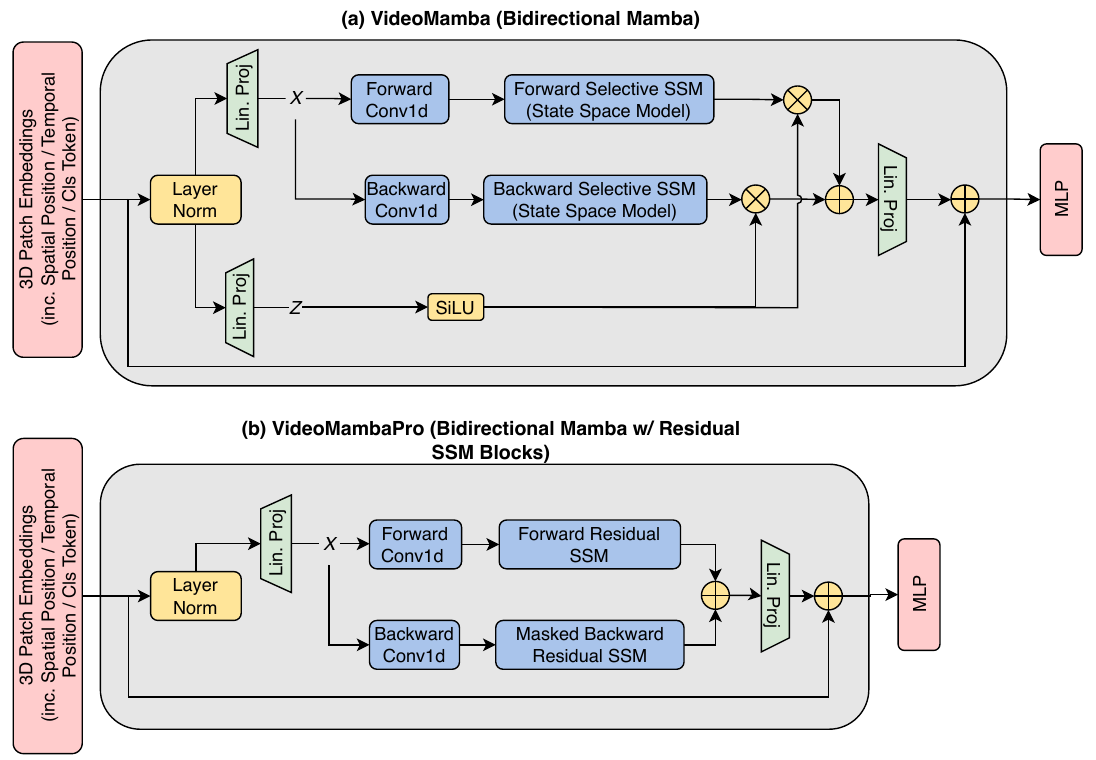}
\caption{(a) VideoMamba Block and (b) VideoMambaPro Architectures. Both employ similar architectures while (a) is the same as \cite{zhu2024vision} adapted for 3D data, (b) uses the concept of residual SSM and masked patches in the backward direction to improve on the architecture in (a).}
\label{fig4}
\end{figure}

The architectures of two VideoMamba models are depicted in Fig.~\ref{fig4}. VideoMamba and its refined variants have been tested against video transformer architectures on tasks such as action recognition, video understanding, and multi-modal tasks \cite{li2025videomamba}. A comparative efficiency analysis between VideoMamba and transformer-based models reveals significant improvements in scalability and performance. \cite{li2025videomamba, lu2024videomambapro, yang2024vivim, arnab2021vivit, chen2024video, bertasius2021space} highlights Mamba's speed advantages when processing videos with a large number of frames, making Mamba a particularly attractive choice for high-volume video processing applications.

\subsection{Comparison}

ViM \cite{zhu2024vision} and VideoMamba \cite{li2025videomamba} both adapt the Mamba state space model (SSM) \cite{gu2023mamba} for visual tasks. ViM serves as a generic vision backbone designed to replace traditional convolutional and attention-based networks \cite{vaswani2017attention}. It processes images as sequences of patches, utilizing bidirectional scanning to efficiently capture spatial context. ViM normalizes input tokens, applies linear projections, and employs a bidirectional SSM with position embeddings to enhance spatial information representation.

VideoMamba extends ViM to 3D video sequences, incorporating multiple scanning methods, including spatial-first, temporal-first, and spatiotemporal scanning. Unlike ViM, it follows a pure SSM-based architecture similar to the vanilla ViT \cite{dosovitskiy2020vit}, without the use of downsampling layers. VideoMamba also introduces a linear-complexity operator to effectively model long-term dependencies, making it particularly suitable for processing spatiotemporal data in video tasks. Both models have been evaluated on tasks like image classification \cite{krizhevsky2012imagenetclassify}, object detection \cite{zhao2019objectdetectreviewdl}, semantic segmentation \cite{xiao2018upernet, yu2018reviewsurveysemseg, hao2020briefsurveysemsegdl}, and many other vision tasks \cite{yang2024vivim, qin2024neuralpalmvein}.

In essence, ViM serves as a general-purpose image processing backbone, while VideoMamba is specifically tailored for efficient video understanding, with architectural adjustments designed to address the unique characteristics and challenges of their respective data types and tasks.

\section{Key Architectures}\label{sec:keyarch}

\begin{figure}[t]
\centering
\includegraphics[width=\linewidth]{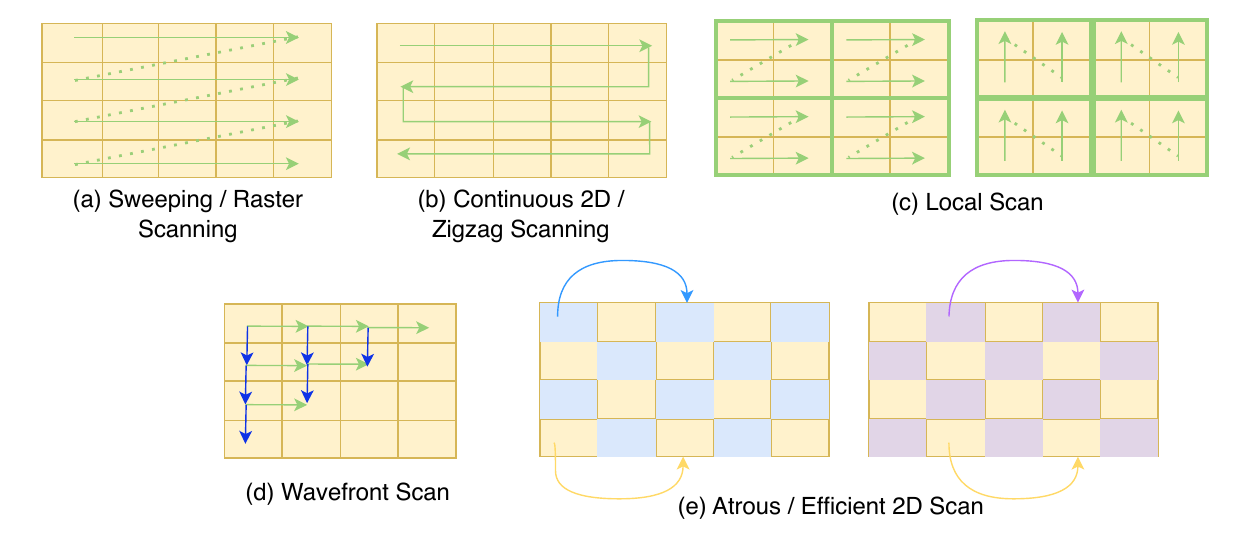}
\caption{Scanning Mechanisms; (a) and (b) are applied on patches in the forward and backward directions, while (a) uses raster scanning, which loses causal information that (b) retains from patches located vertically from each other. Local Scan (c) divides an image into local segments for improved local representation. Scans shown in (d) and (e) are not commonly used scanning mechanisms but are efficient alternatives to (a) and (b).}
\label{fig5}
\end{figure}

This section summarizes the recent advancements in Vision/Video Mamba-based studies from the perspectives of key architectural innovations. Fig.~\ref{fig5} shows additional selective scan mechanisms employed by the architecture variants described below. Each novel scanning strategy has been configured to supplement using Mamba for 2D and 3D applications, such as spatial relationship modeling, limited local representations, and varying directional scanning for better visual contextual features.  Fig.~\ref{figfull} shows visual block representations of some of the key architectures used in image tasks and how these blocks compare to the original Mamba block in Fig.~\ref{fig1}.

\paragraph{Selective State Space Models (SSMs)} This is the foundation of the Mamba architecture. SSMs, especially the S6 block, offer a global receptive field and linear complexity in relation to sequence length, making them a computationally efficient alternative to traditional Transformers \cite{gu2023mamba, zou2024venturing, shi2024vssd, lee2024efficientvim}. Mamba introduces a selection mechanism, enabling the model to selectively retain or discard information based on the input, enhancing context-based reasoning \cite{gu2023mamba, zou2024venturing}. This architecture enables the model to perform content-based reasoning by making the SSM parameters functions of the input. This allows the models to selectively propagate or forget information along the sequence length, filtering irrelevant information and maintaining long-term memory of relevant information \cite{gu2023mamba, zou2024venturing, dao2024mamba2, you2024mambabev}.
\paragraph{Multi-path Scanning Mechanisms} To overcome the limitations of Mamba's causal nature, which is not ideal for image processing, several models like VMamba in Fig.\ref{figfull}(b), LocalMamba in Fig.\ref{figfull}(c), and PlainMamba in Fig.\ref{figfull}(d) have adopted multi-path scanning mechanisms. This allows the models to capture both local and global features more effectively \cite{zhu2024vision, yang2024plainmamba, huang2024localmamba}.

LocalMamba \cite{huang2024localmamba} improves image scanning by dividing images into distinct local windows, ensuring relevant tokens are closely arranged to better capture local dependencies while maintaining spatial coherence. To incorporate global context, the method integrates a selective scan mechanism across four directions: the original two and their flipped counterparts, allowing for both forward and backward token processing, as shown in Fig.~\ref{fig5}(c).

\paragraph{Structure-Aware SSM Layer} Introduced in Spatial-Mamba, this layer incorporates a Structure-Aware State Fusion (SASF) branch and a multiplicative gate branch as shown in Fig.~\ref{figfull}(e). This allows the model to capture both local and global spatial dependencies \cite{xiao2024spatial}.
Unlike previous methods that rely on multiple scanning directions, Spatial-Mamba fuses the state variables into a new structure-aware state variable \cite{xiao2024spatial}. Once an input image is flattened into 1D patches and state parameters are determined, the patches are reshaped into 2D, and SASF is applied to integrate local dependencies before generating the final output.

\paragraph{Cross-Layer Token Fusion} Famba-V, a technique designed to enhance the training efficiency of Vision Mamba (ViM) models, incorporates three cross-layer token fusion strategies \cite{shen2024famba, ding2022towardsquantvit, bolya2022tokenmerge, cao2023pumertokenmerge}.

Token fusion is applied within the ViM block as shown before the final linear projection operation in Fig.~\ref{figfull} (f), the tokens are grouped and paired based on cosine similarity, and fused by averaging. There are three token fusion strategies implemented in \cite{shen2024famba} with upper layer token fusion, producing the best results on vision tasks shown in Section V:

\begin{itemize}
    \item Interleaved Token Fusion -- Applies fusion to alternate layers, starting from the second layer, to maintain efficiency without overly aggressive fusion.
    \item Lower-layer Token Fusion -- Applies fusion only to the lower layers, leveraging early-stage efficiency gains while preserving high-level reasoning.
    \item Upper-layer Token Fusion -- Applies fusion only to upper layers, assuming high-level features can be fused without significantly affecting performance.
\end{itemize}

\paragraph{Hierarchical Mamba Architecture (Hi-Mamba)} This architecture in Fig.~\ref{figfull}(g) designed for image super-resolution, introduces a hierarchical Mamba block (HMB) comprised of local and region SSMs with single-direction scanning. This setup aims to capture multi-scale visual context efficiently \cite{qiao2024hi}.

\paragraph{Direction Alternation Hierarchical Mamba Group (DA-HMG)} Within Hi-Mamba, this group further enhances spatial relationship modeling. It allocates single-direction scanning into cascaded HMBs, improving performance without increasing computational costs or parameters \cite{qiao2024hi}.

\paragraph{Non-Causal State Space Duality (NC-SSD)} Models like VSSD introduce non-causal SSD, eliminating the causal mask in the state space dual model (SSD) of Mamba2 \cite{dao2024mamba2}. This architectural innovation allows for a more flexible application of Mamba to inherently non-causal data like images \cite{zhu2024vision, shi2024vssd, gu2021efficiently, lee2024efficientvim}.

The VSSD block \cite{shi2024vssd} enhances Mamba2 SSD for vision tasks by incorporating Depth-Wise Convolution (DWConv) \cite{he2017maskrcnn}, a Feed-Forward Network (FFN) for better channel communication, and a Local Perception Unit (LPU) for improved local feature extraction. The four-stage hierarchical VSSD model uses VSSD blocks in the first three stages and Multi-Head Self-Attention (MSA) in the last, balancing efficiency and performance in vision applications \cite{vaswani2017attention}.

\paragraph{Hidden State Mixer-based SSD (HSM-SSD)} EfficientViM models feature this layer, which shifts the channel mixing operations from the image feature space to the hidden state space. This design aims to alleviate the computational bottleneck of the standard SSD layer without compromising the model's ability to generalize \cite{lee2024efficientvim}.

\paragraph{Register Tokens in Vision Mamba} Mamba® introduces registers, input-independent tokens, into the Vision Mamba architecture. These tokens, inserted evenly throughout the input and recycled for final predictions, aim to mitigate artifacts and enhance the model's focus on semantically meaningful image regions \cite{wang2024mambareg}.

\paragraph{Masked Backward Computation} VideoMambaPro incorporates masked backward computation during the bidirectional Mamba process, mitigating the issue of historical decay in token processing and enabling the model to better utilize historical information \cite{li2025videomamba, lu2024videomambapro}.
\paragraph{Elemental Residual Connections} VideoMambaPro also introduces residual connections to the matrix elements of Mamba, addressing the challenge of element contradiction, where elements in the sequence can conflict during computation. This enhances the model's ability to extract and process complex spatio-temporal features in video data \cite{li2025videomamba, lu2024videomambapro}.

\begin{comment}
\paragraph{Other Architectures combining other Vision Architectures}
\paragraph{Multi-head Mamba Module} The Global-local Vision Mamba (GLVM) uses a multi-head Mamba module with a multi-direction scanning mechanism. It also integrates branches for CNN, MHMamba, and a Feature Interaction Unit to capture both local and global representations for tasks like vein recognition \cite{qin2024neuralpalmvein}.

\paragraph{Modulated Group Mamba Layer} This layer within the Group Mamba architecture was introduced to address limitations of standard vision Mamba blocks, aims to improve stability and manage the number of parameters as the number of channels increases\cite{shaker2024groupmamba}.
\end{comment}

%\vspace{12pt}
Many of these innovations focus on enhancing the efficiency and performance of Mamba-based models in vision tasks, while preserving the linear complexity advantage over traditional Transformers.

\section{Performance Comparison}
\label{sec:results}
To provide a systematic comparison of the performance of different models across various datasets and tasks, this section compares the results of \cite{gu2023mamba, zhu2024vision, liu2024vmambavisualstatespace, xiao2024spatial, shen2024famba, huang2024localmamba, shi2024vssd, dao2024mamba2, lee2024efficientvim, yang2024plainmamba, wang2024mambareg, li2025videomamba, lu2024videomambapro} in common dataset benchmarks. The models shown are only those of a similar size in the cases where researchers provided `small', `base', or `large' model variations. The results are categorized by tasks; image classification against the Imagenet-1k dataset \cite{deng2009imagenetdb, krizhevsky2012imagenetclassify}, semantic segmentation against the ADE20K dataset \cite{yu2018reviewsurveysemseg, zhou2019semanticade20k}, object detection against the MS-COCO dataset \cite{lin2014microsoftcocodataset}, and human action recognition for video understanding benchmarks against both Kinetics-400 \cite{kay2017kinetics, carreira2017quo} and Something-Something-V2 \cite{goyal2017ssv2} datasets.

\begin{table*}[t]
    \centering
    \small
    \caption{Comparison of Different Image Classification Model Results}
    \resizebox{\textwidth}{!}{
    \begin{tabular}{|l|c|c|l|c|c|c|}
        \hline
        \textbf{Model} & \textbf{Data} & \textbf{Task} & \textbf{Scanning Mechanism} & \textbf{\#Params (M)} & \textbf{FLOPS (G)} & \textbf{Top-1 ACC} \\
        \hline
        Vision Mamba (ViM-S) \cite{zhu2024vision} & ImageNet-1k & Image Classification & Bidirectional Scan (Sweeping / Raster Scan) & 26 & - & 80.3 \\
        Vmamba-S \cite{liu2024vmambavisualstatespace} & ImageNet-1k & Image Classification & Cross-Scan (Raster Scanning along 4 directions) & 50 & 8.7 & 83.6 \\
        Spatial Mamba-S \cite{xiao2024spatial} & ImageNet-1k & Image Classification & Sweeping Scan with Structure Aware State Fusion & 43 & 7.1 & 84.6 \\
        Famba-V applied to ViM-S \cite{shen2024famba} & CIFAR-100 & Image Classification & Bidirectional Scan w/ Token Fusion & 26 & - & 75.2 \\
        LocalMamba applied to ViM-S \cite{huang2024localmamba} & ImageNet-1k & Image Classification & Local Scan (Sweeping/ Raster Scan) & 28 & 4.8 & 81.2 \\
        LocalMamba applied to Vmamba-S \cite{huang2024localmamba} & ImageNet-1k & Image Classification & Local Scan (Sweeping/ Raster Scan) & 50 & 11.4 & 83.7 \\
        VSSD-S (based on Mamba2) \cite{shi2024vssd} & ImageNet-1k & Image Classification & Bidirectional Scan w/ Non-Causal State Space Duality & 40 & 7.4 & 84.1 \\
        EfficientViM-M4 \cite{lee2024efficientvim} & ImageNet-1k & Image Classification & Bidirectional Scan w/ Hidden State Mixer SSD & 21.3 & 4.1 & 81.9 \\
        PlainMamba-L2 \cite{yang2024plainmamba} & ImageNet-1k & Image Classification & Continuous 2D / Zigzag Scan & 25.7 & 8.1 & 81.6 \\
        Mamba®-S \cite{wang2024mambareg} & ImageNet-1k & Image Classification & Bidirectional Scan w/ evenly distributed Register Token Embeddings & 28 & 9.9 & 81.1 \\
        \hline
    \end{tabular}
    }
    \label{tab:image_classification}
\end{table*}
\begin{table*}[t]
    \centering
    \small
    \caption{Comparison of Different Semantic Segmentation Model Results}
    \resizebox{\textwidth}{!}{
    \begin{tabular}{|l|c|c|l|c|c|c|}
        \hline
        \textbf{Backbone with Upernet Model} & \textbf{Data} & \textbf{Task} & \textbf{Scanning Mechanism} & \textbf{\#Params (M)} & \textbf{FLOPS (G)} & \textbf{mIoU} \\
        \hline
        Vision Mamba (ViM-S) \cite{zhu2024vision} & ADE20K & Semantic Segmentation & Bidirectional Scan (Sweeping / Raster Scan) & 46 & - & 44.9 \\
        Vmamba-S \cite{liu2024vmambavisualstatespace} & ADE20K & Semantic Segmentation & Cross-Scan (Raster Scanning along 4 directions) & 82 & 1028 & 50.6 \\
        Spatial Mamba-S \cite{xiao2024spatial} & ADE20K & Semantic Segmentation & Sweeping Scan with Structure Aware State Fusion & 73 & 992 & 50.6 \\
        LocalMamba applied to ViM-S \cite{huang2024localmamba} & ADE20K & Semantic Segmentation & Local Scan (Sweeping/ Raster Scan) & 58 & 297 & 46.4 \\
        LocalMamba applied to Vmamba-S \cite{huang2024localmamba} & ADE20K & Semantic Segmentation & Local Scan (Sweeping/ Raster Scan) & 81 & 1095 & 50 \\
        VSSD-T (based on Mamba2) \cite{shi2024vssd} & ADE20K & Semantic Segmentation & Bidirectional Scan w/ Non-Causal State Space Duality & 53 & 941 & 47.9 \\
        PlainMamba-L2 \cite{yang2024plainmamba} & ADE20K & Semantic Segmentation & Continuous 2D / Zigzag Scan & 55 & 285 & 46.8 \\
        Mamba® \cite{wang2024mambareg} & ADE20K & Semantic Segmentation & Bidirectional Scan w/ evenly distributed Register Token Embeddings & 56 & - & 45.3 \\
        \hline
    \end{tabular}
    }
    \label{tab:semantic_segmentation}
\end{table*}

\subsection{Image Classification}

Table ~\ref{tab:image_classification} compares various Mamba models used for image classification \cite{krizhevsky2012imagenetclassify, he2016deepresimagerec} on the ImageNet-1k dataset \cite{deng2009imagenetdb}, emphasizing differences in scanning mechanisms, accuracy, and efficiency. From the results we can see that Spatial-Mamba-S (84.6\% Top-1 ACC) \cite{xiao2024spatial} is the best performer, leveraging Structure-Aware State Fusion, while Vmamba-S (83.6\%) \cite{liu2024vmambavisualstatespace} and LocalMamba applied to Vmamba-S (83.7\%) \cite{huang2024localmamba, zhu2024vision} also achieve strong results using cross-scan and local scan. Famba-V applied to ViM-S (75.2\%) \cite{shen2024famba, zhu2024vision} performs the worst, although tested on a different dataset, suggests bidirectional scanning with token fusion is less effective for image classification. In terms of efficiency, EfficientViM-M4 (21.3M params, 4.1 GFLOPS, 81.9\% ACC) \cite{lee2024efficientvim} provides an excellent accuracy-to-computation trade-off, making it ideal for resource-constrained applications. 

The results show that advanced state-space fusion (Spatial-Mamba-S) and strategic scanning mechanisms (Vmamba-S, LocalMamba) significantly enhance accuracy, while lightweight architectures (EfficientViM-M4) \cite{lee2024efficientvim} can balance efficiency and accuracy effectively.

\subsection{Semantic Segmentation}

Table ~\ref{tab:semantic_segmentation} compares the reviewed Mamba models on the task of semantic segmentation on ADE20K \cite{zhou2019semanticade20k, xiao2018upernet, yu2018reviewsurveysemseg, hao2020briefsurveysemsegdl}. The results show that 
Vmamba-S \cite{liu2024vmambavisualstatespace} and Spatial-Mamba-S (50.6 mIoU) \cite{xiao2024spatial} achieve the best performance, leveraging cross-scan and Structure-Aware State Fusion, though they demand high computational costs (1028G and 992G FLOPS). LocalMamba applied to Vmamba-S (50.0 mIoU, 1095G FLOPS) also performs well \cite{huang2024localmamba}, emphasizing local scanning benefits. MambaR-S (45.3 mIoU) performs the worst, showing register token embeddings are less effective for segmentation \cite{wang2024mambareg}. VSSD-T (47.9 mIoU), although only tested with a `tiny' version of the model, balances accuracy and efficiency with non-causal state space duality \cite{shi2024vssd}. 

The results highlight the trade-offs between scanning mechanisms, mIoU, and efficiency (Params and FLOPS). This analysis indicates that high-performing models depend on advanced scanning techniques but demand considerable computational resources, while lighter models, such as PlainMamba-L2 \cite{yang2024plainmamba}, sacrifice some accuracy for improved efficiency.

\subsection{Object Detection}

Table ~\ref{tab:object_detection} evaluates object detection models \cite{zhao2019objectdetectreviewdl} on MS-COCO \cite{lin2014microsoftcocodataset, he2017maskrcnn}, comparing accuracy based on 75\% overlapping bounding box with ground truth labels (AP\textsubscript{bb}\textsuperscript{75}) and efficiency (Params and FLOPS). 

Spatial Mamba-S (54.2 AP\textsubscript{bb}\textsuperscript{75}) is the best performer, benefiting from Sweeping Scan with Structure-Aware State Fusion, although it has high computational demands (315G FLOPS) \cite{xiao2024spatial}. VSSD-S (53.1 AP\textsubscript{bb}\textsuperscript{75}) follows closely, leveraging Non-Causal State Space Duality, balancing accuracy and efficiency (325G FLOPS) \cite{shi2024vssd}. LocalMamba applied to Vmamba-S (52.7 AP\textsubscript{bb}\textsuperscript{75}) also performs well, emphasizing local scanning advantages \cite{huang2024localmamba}. The worst performer is EfficientViM-M4 (41.1 AP\textsubscript{bb}\textsuperscript{75}), which, despite its efficiency (4.1G FLOPS), sacrifices accuracy in place of reduced computational cost \cite{lee2024efficientvim}. Vmamba-S (48.7 AP\textsubscript{bb}\textsuperscript{75}) and Vision Mamba (49.6 AP\textsubscript{bb}\textsuperscript{75}) \cite{liu2024vmambavisualstatespace, zhu2024vision}, which rely on cross-scan and bidirectional raster scanning, show moderate performance but are less effective than structure-aware or local scanning methods. PlainMamba-L2 (50.1 AP\textsubscript{bb}\textsuperscript{75}, 542G FLOPS) \cite{yang2024plainmamba} achieves decent results but is computationally expensive. This analysis highlights that Spatial Mamba-S and VSSD-S lead in accuracy, while EfficientViM-M4 prioritizes efficiency at the cost of performance.

\begin{table*}[t]
    \centering
    \small
    \caption{Comparison of Different Object Detection Model Results}
    \resizebox{\textwidth}{!}{
    \begin{tabular}{|l|c|c|l|c|c|c|}
        \hline
        \textbf{Backbone with Mask R-CNN Model} & \textbf{Data} & \textbf{Task} & \textbf{Scanning Mechanism} & \textbf{\#Params (M)} & \textbf{FLOPS (G)} & \textbf{AP$^{box}_{75}$} \\
        \hline
        Vision Mamba (ViM-Ti) \cite{zhu2024vision} & MS-COCO & Object Detection & Bidirectional Scan (Sweeping / Raster Scan) & 49.6 & - & - \\
        Vmamba-S \cite{liu2024vmambavisualstatespace} & MS-COCO & Object Detection & Cross-Scan (Raster Scanning along 4 directions) & 70 & 349 & 48.7 \\
        Spatial Mamba-S \cite{xiao2024spatial} & MS-COCO & Object Detection & Sweeping Scan with Structure Aware State Fusion & 63 & 315 & 54.2 \\
        LocalMamba applied to Vmamba-S \cite{huang2024localmamba} & MS-COCO & Object Detection & Local Scan (Sweeping/ Raster Scan) & 69 & 414 & 52.7 \\
        VSSD-S (based on Mamba2) \cite{shi2024vssd} & MS-COCO & Object Detection & Bidirectional Scan w/ Non-Causal State Space Duality & 59 & 325 & 53.1 \\
        EfficientViM-M4 \cite{lee2024efficientvim} & MS-COCO & Object Detection & Bidirectional Scan w/ Hidden State Mixer SSD & 21.3 & 4.1 & 41.1 \\
        PlainMamba-L2 \cite{yang2024plainmamba} & MS-COCO & Object Detection & Continuous 2D / Zigzag Scan & 53 & 542 & 50.1 \\
        \hline
    \end{tabular}
    }    
    \label{tab:object_detection}
\end{table*}

\begin{table*}[t]
    \centering
    \small
    \caption{Comparison of Different Video Understanding -- Human Action Recognition Model Results}
    \resizebox{\textwidth}{!}{
    \begin{tabular}{|l|c|c|l|c|c|c|}
        \hline
        \textbf{Backbone/Model} & \textbf{Data} & \textbf{Task} & \textbf{Scanning Mechanism} & \textbf{\#Params (M)} & \textbf{FLOPS (G)} & \textbf{Top-1 ACC} \\
        \hline
        VideoMamba-S \cite{li2025videomamba} & Kinetics-400 & Action Recognition & 3D Bidirectional Scan & 26 & 1620 & 81.5 \\
        VideoMamba-S \cite{li2025videomamba} & SSv2 & Action Recognition & 3D Bidirectional Scan & 26 & 408 & 67.6 \\
        VideoMambaPro-S \cite{lu2024videomambapro} & Kinetics-400 & Action Recognition & 3D Bidirectional Scan w/ Residual SSM + Masked Backward Residual SSM & 24 & 1500 & 88.5 \\
        VideoMambaPro-S \cite{lu2024videomambapro} & SSv2 & Action Recognition & 3D Bidirectional Scan w/ Residual SSM + Masked Backward Residual SSM & 24 & 400 & 74.3 \\
        \hline
    \end{tabular}
    }
    \label{tab:video_understanding}
\end{table*}

\subsection{Human Action Recognition}

Table~\ref{tab:video_understanding} compares the VideoMamba-S \cite{li2025videomamba} and VideoMambaPro-S \cite{lu2024videomambapro} models tested on the Kinetics-400 \cite{kay2017kinetics} and SSV2 \cite{goyal2017ssv2} datasets for action recognition. VideoMambaPro-S emerges as the superior model, achieving the highest accuracy of 88.5\% on Kinetics-400 while using fewer parameters (24M vs 26M) than VideoMamba-S. Both models consistently perform better on Kinetics-400 than SSV2. Ref. \cite{lu2024videomambapro} utilizes an enhanced scanning mechanism, which includes additional Residual SSM and Masked Backward Residual SSM components \cite{lu2024videomambapro, he2017maskrcnn, cheng2022maskedattntrans}, which contributed to better performance across both datasets. Although computational costs (FLOPS) are significantly lower for SSV2 experiments (400G) compared to Kinetics-400 (1500-1620G), the worst-performing configuration is VideoMamba-S in SSV2 with 67. 6\% accuracy. Overall, the Pro version demonstrates that its more sophisticated architecture delivers better results with slightly improved efficiency.

\subsection{Discussion}\label{sec:discuss}

%In this section, we discuss our recommendation for which models the reader should use for specific vision tasks, and which models would produce worse performing results.
Based on the above experimental evaluations, it is evident that Spatial Mamba-S and Vmamba-S are the best choices for image classification (84.6\% and 83.6\% Top-1 ACC) and semantic segmentation (50.6 mIoU), though they require high computational cost. For object detection, Spatial Mamba-S (54.2 AP\textsubscript{bb}\textsuperscript{75}) and VSSD-S (53.1 AP\textsubscript{bb}\textsuperscript{75}) perform best, leveraging structure-aware and non-causal mechanisms. VSSD-S and LocalMamba offer a balance between accuracy and efficiency across tasks. %Models like EfficientViM-M4 (41.1 AP\textsubscript{bb}\textsuperscript{75}) and Mamba®-S (45.3 mIoU) underperform and should be avoided, while PlainMamba-L2 is moderate but lacks standout results.

\section{Challenges and Future Research}\label{sec:chall}

Mamba models are capable of modeling long-range dependencies with linear complexity, making them well-suited for tasks involving high-resolution images, long sequences, and large video datasets. Techniques such as bidirectional scanning and multi-scale processing further enhance Mamba's ability to capture both local and global context. Mamba is currently being explored in specialized vision domains, such as:

\begin{itemize}
    \item Scene flow estimation. FlowMamba \cite{lin2024flowmamba} uses an iterative SSM-based update module and a feature-induced ordering strategy to capture long-range motion.
    \item Vein recognition. Global-local Vision Mamba (GLVM) combines local and global feature learning for enhanced vein recognition \cite{qin2024neuralpalmvein}.
    \item Medical image analysis. PV-SSM is a pure visual state space model achieving strong results across medical image analysis \cite{wang2024pvssm}.
    \item Skin lesion segmentation. MambaU-Lite \cite{nguyen2024mambaulite} combines Mamba and CNN architectures for efficient skin lesion segmentation, using a novel P-Mamba block.
\end{itemize}

Future research aims to develop native 2D and 3D Mamba models, non-causal state-space dual models, and efficient token fusion techniques. Some potential directions include:

\begin{itemize}
    \item Multi-modal image fusion. Shuffle Mamba employs a random shuffle-based scanning method to mitigate biases in multi-modal image fusion \cite{cao2024shuffle}.
    \item RGB-D salient object detection. MambaSOD is a dual Mamba-driven network for RGB-D salient object detection, using cross-modal fusion to combine RGB and depth information \cite{zhan2024mambasod}.
    \item Point cloud processing. NIMBA reorders point cloud data to maintain 3D spatial structure, enhancing Mamba's sequential processing \cite{koprucu2024nimba}. Serialized Point Mamba \cite{wang2024serializedpointcloudmamba} uses a state-space model for dynamic compression of point cloud sequences for efficient segmentation.
\end{itemize}

Combining Mamba with CNNs and Transformers is also a research direction to compensate for each other's weaknesses.

\begin{itemize}
    \item Object detection with YOLO. Mamba-YOLO integrates Mamba into the YOLO architecture, achieving improved object detection performance \cite{wang2024mambayolo}.
    \item 3D object detection. MambaBEV \cite{you2024mambabev} and PillarMamba \cite{liu2024pillarmamba} integrate Mamba into 3D object detection models for autonomous driving. MambaDETR is also used for 3D object detection \cite{ning2024mambadetr}.
    \item Point cloud enhancement. MambaTron uses Mamba for cross-modal point cloud completion, combining Mamba's efficiency with Transformer's analytical capabilities \cite{inaganti2025mambatron}.
\end{itemize}

VideoMamba \cite{li2025videomamba} still has challenges in achieving performance parity with Transformers, addressing historical decay, and managing potential element contradictions \cite{sinha2025mstemba}. Future research will focus on enhancing stability and closing the performance gap with other architectures, in order to fully unlock Mamba's potential for efficient, long-range context modeling in vision and video.

\section{Conclusion}\label{sec:concl}

This paper provides an overview of visual Mamba, detailing their architectures, applications, and challenges. Vision Mamba functions as a general vision backbone, processing images as sequences of patches with bidirectional scanning to effectively capture spatial context. Video Mamba extends this approach to 3D video sequences, operating across both spatial and temporal dimensions.
Despite significant advancements and key architectural developments, challenges persist in fully realizing Mamba’s potential for vision tasks. Future research should focus on refining these architectures, broadening Mamba’s applicability to a wider range of visual modalities, developing efficient token fusion techniques, and improving its scalability for real-world deployment.

\section*{Acknowledgment}
This research was funded by the Natural Sciences and Engineering Research Council of Canada (NSERC) under grant numbers ALLRP 588173-23 and ALLRP 570580-21.
%--------------------------------------------------------------------------%
%\clearpage

%\clearpage

% References
\balance
% Specify the bibliography style
\bibliographystyle{IEEEtran}
% Specify the bibliography file
\bibliography{Reference/references}

\vspace{12pt}
\end{document}